\documentclass[journal]{IEEEtran}
\usepackage{kotex}
\usepackage{hyperref}
\usepackage{multirow}
\usepackage{multicol}
\usepackage{subcaption}
\usepackage{hyperref}
\usepackage{xcolor}
\hypersetup{
    colorlinks,
    linkcolor={red!90!black},
    citecolor={blue!90!black},
    urlcolor={blue!90!black}
}
\usepackage[pdftex]{graphicx}
\graphicspath{{../pdf/}{../jpeg/}{../png/}}
\DeclareGraphicsExtensions{.pdf,.jpeg,.png}
\usepackage{array, makecell, pbox, bigstrut}
\usepackage{booktabs} 
\usepackage{amsmath}

\ifCLASSINFOpdf
\else
\fi
\begin{document}
%
\title{ Content-Based Video\textendash Music Retrieval \\ Using Soft Intra-Modal Structure Constraint
}
%
%
%

\author{Sungeun~Hong,~\IEEEmembership{Student Member,~IEEE,}
        Woobin~Im, 
        and Hyun~S.~Yang,~\IEEEmembership{Member,~IEEE}
}

\maketitle

\begin{abstract}
Up to now, only limited research has been conducted on cross-modal retrieval of suitable music for a specified video or vice versa. 
Moreover, much of the existing research relies on metadata such as keywords, tags, or associated description that must be individually produced and attached posterior. 
This paper introduces a new content-based, cross-modal retrieval method for video and music that is implemented through deep neural networks. 
We train the network via  inter-modal ranking loss such that videos and music with similar semantics end up close together in the embedding space. 
However, if only the inter-modal ranking constraint is used for embedding,   modality-specific characteristics can be lost.
To address this problem, we propose a novel soft intra-modal structure loss that leverages the relative distance relationship between intra-modal samples before embedding.
We also introduce reasonable quantitative and qualitative experimental protocols to solve the lack of  standard protocols for less-mature video\textendash music related tasks.
Finally, we construct  a large-scale 200K video\textendash music pair benchmark.
All the datasets and source code can be found in our online repository (\href{https://github.com/csehong/VM-NET}{https://github.com/csehong/VM-NET}).
\end{abstract}


%
\IEEEpeerreviewmaketitle

\begin{figure*}
    \centering
    \begin{subfigure}[t]{0.5\textwidth}
        \centering
        \includegraphics[height=1.65in]{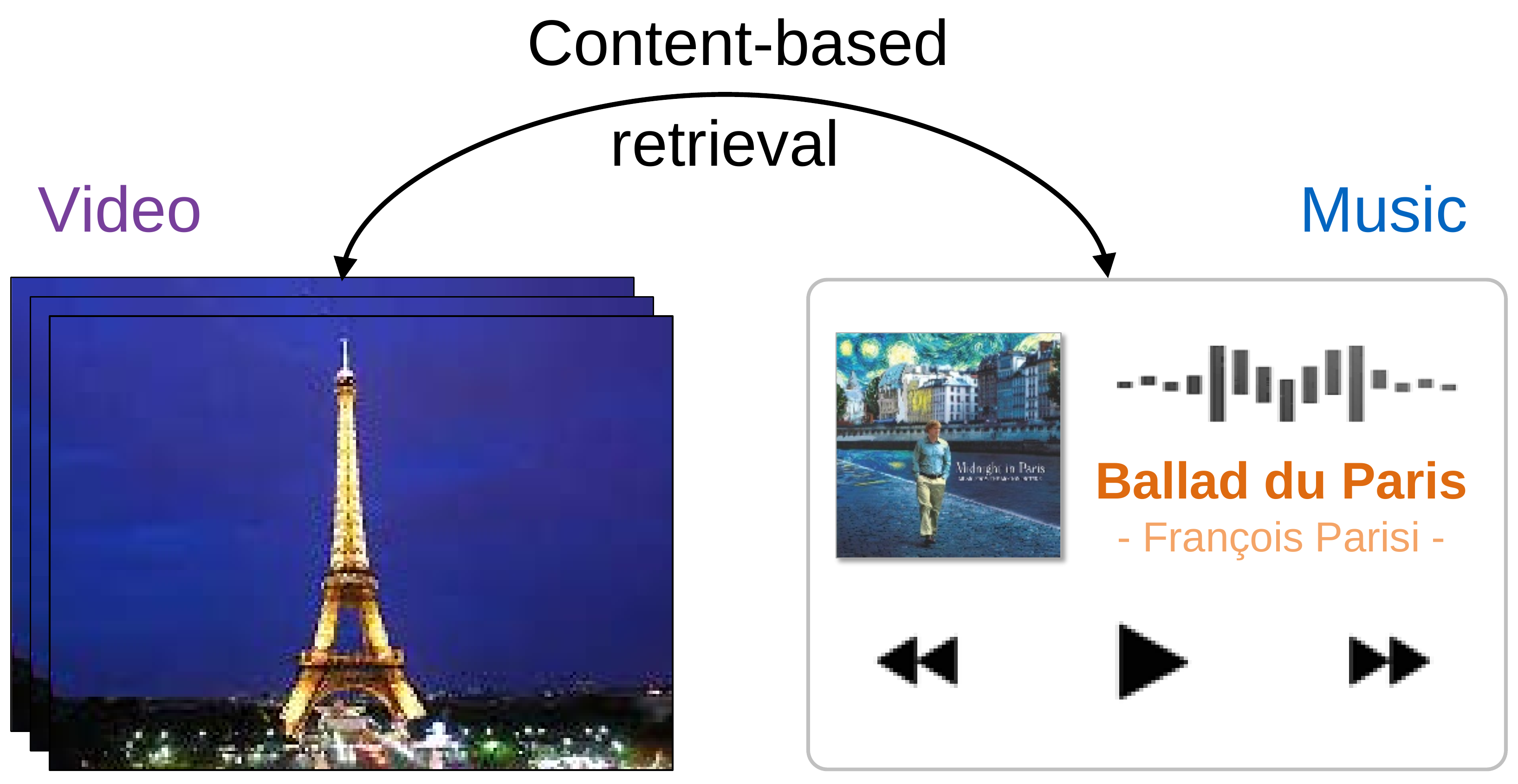}
        \caption{Eiffel Tower scenes $\Longleftrightarrow$ Music based on the Paris night view}
        \label{fig:eiffel}
    \end{subfigure}%
    ~ 
    \begin{subfigure}[t]{0.5\textwidth}
        \centering
        \includegraphics[height=1.65in]{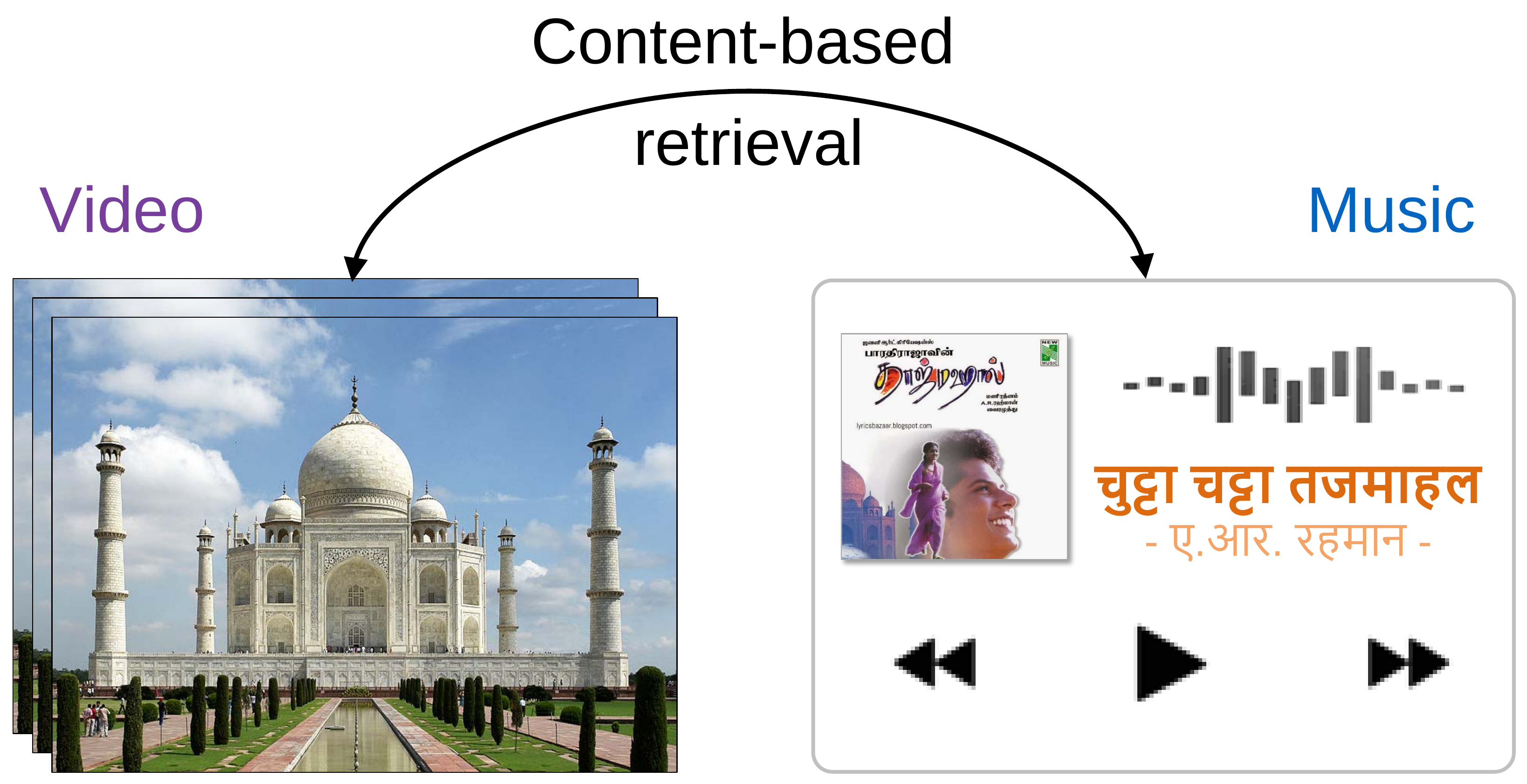}
        \caption{Taj Mahal scenes $\Longleftrightarrow$ Traditional Indian music}
        \label{fig:tajmahal}
    \end{subfigure}
    \caption{Content-based video\textendash music retrieval (\textbf{CBVMR}):  
    Unlike most of existing methods  use metadata (e.g., keywords, mood tags, and associated description) to associate visual modalities with music, our model performs bidirectional retrieval  tasks between video and music using image frames for the video parts and audio signals for the music parts.  }
    \label{fig:motivation}
\end{figure*}

\section{Introduction}
Music, images, and videos are widely used in everyday life and as a result, research in this field has been actively conducted over recent decades. However, to date, the relation between music and visual modalities (e.g., image or video) has not been adequately investigated. Previous studies have focused primarily on how data of a single modality might be processed in various tasks, e.g., music genre classification, music information retrieval, melody extraction, image retrieval, and video classification \cite{mayer2008combination, kum2016melody, kim2009using, szegedy2015going, wan2014deep}. Some pioneering approaches that have explored the relation between music and visual modalities have been developed \cite{brochu2003sound, mayer2011analysing, chao2011tunesensor, acar2014understanding, wu2016bridging}. However, these techniques mainly use metadata, which is separately attached to each individual item of music or visual data, rather than content-based information that can be derived directly from the music or visual data itself.

We argue that content-based approaches are more appropriate than metadata-based approaches for several reasons. First, metadata-based systems require  writing the meta-information for each data item individually, and this can be impractical for large-scale datasets. Moreover, in this situation, as mentioned in \cite{van2013deep}, unpopular multimedia items can be neglected because their meta-information is scarce or difficult to obtain. 
Second, metadata-based approaches often rely on a hard-coded mapping function that links concepts from a visual modality to music templates, and this imposes limitations on their various applications. For example, to generate soundtracks for user-generated videos, Yu et al. \cite{yu2012automatic} mapped a video's geometric meta-information to a mood tag and then recommended music with the corresponding mood tag. 

In this study, we undertook the task of finding music that suits a given video and vice versa using a content-based approach, which can be used for bidirectional retrieval as shown in Fig.~\ref{fig:motivation}. 
The main challenge in this task  is to design a cross-modal model that has no requirements for metadata and, additionally, has no hard-coded mapping functions or assumptions about specific rules.
The second, practical challenge is  the difficulty of obtaining well-matched video\textendash music pairs  that are essential for data-driven learning such as deep learning.
The matching criteria between video and music  are more ambiguous than that of other cross-modal tasks such as image-to-text retrieval; as a result, there are very few public datasets that associate visual modalities and music.

Our key insight is that if the video\textendash music pair dataset is obtainable, then the relationship between video and music can be learned through  deep neural networks. 
To obtain video\textendash music pairs, we leverage large-scale music video datasets by treating each audio and visual signal from the same music video as a ground truth pair. 
This is motivated by studies showing that customers usually use album cover art as a visual cue when browsing music in record stores \cite{mayer2011analysing}. We are also inspired by the fact that producers with professional skill carefully create each album cover or music video considering the singer and the characteristics of the  songs \cite{wu2016bridging}. 
Concretely, the contributions of this research are three-fold:

1) \textbf{Content-based, cross-modal  embedding network}. 
We introduce  VM-NET, two-branch neural network that infers the latent alignment between videos and music tracks using only their contents.
We train the network via inter-modal ranking loss, such that videos and music with similar semantics end up close together in the embedding space. 
However, if only the inter-modal ranking constraint for embedding is considered, modality-specific characteristics (e.g., rhythm or tempo for music and texture or color for image) may be lost.
To solve this problem, we devise a novel soft intra-modal structure constraint that takes advantage of the relative distance relationship of samples within each modality.
Unlike conventional approaches, our model does not require ground truth pair information within individual modality.

2) \textbf{Large-scale video\textendash music pair dataset}. Owing to the lack of video\textendash music datasets, we have compiled a large-scale benchmark called Hong\textendash Im Music\textendash Video 200K (HIMV-200K) composed of 200,500 video\textendash music pairs.
While most previous studies have used only official music videos, which limit the musical style or amount of data, our HIMV-200K includes official music videos, parody music videos, and user-generated videos.

3) \textbf{Reasonable experimental protocols}. 
Existing video\textendash music related approaches have few standard protocols for evaluation, and even  most of them mainly performed subjective user evaluation.
For quantitative evaluation, we use Recall@$K$, which can give a baseline for subsequent content-based video\textendash music retrieval ({CBVMR}) studies. 
We also suggest and conduct a reasonable subjective test that examines user preferences for videos that suit given music and, vice versa.  
Finally, we present a qualitative investigation of the retrieval results  using our model.

\section{Related Work}
\subsection{Video\textendash Music Related Tasks}
Early investigations of the relationship between music and visual modalities have involved studies using album covers as the visual modality \cite{brochu2003sound, mayer2011analysing, libeks2011you, chao2011tunesensor}.
As online music streaming (e.g., Apple Music, Pandora, Last.fm) and video sharing sites (e.g., YouTube, Vimeo, Daily motion) have become popular, recent related research has included studies using music videos \cite{gillet2007correlation, acar2014understanding, schindler2015audio, wu2016bridging} instead of static album covers. 
Overall, conventional approaches can be divided into three categories according to the task: generation, classification, and matching.

The earlier studies of generation tasks include automatic music video generation.
Hua et al. \cite{hua2004automatic}  suggested a system that generates music videos  from personal  videos by using  temporal structure and repetitive patterns.
Shamma et al. \cite{shamma2005musicstory} introduced MusicStory, a music video creator that builds a video for a piece of music using relevant images.
To this end, they processed the lyrics in the music, which retrieves related images from photo-sharing websites.
The correlations of the audio and visual signals in music videos has also been investigated \cite{gillet2007correlation}. 

For music genre classification, Libeks et al \cite{libeks2011you} proposed a framework that use promotional photos of artists as well as album covers.
They predict music genre tags based on content-based image analysis using color and texture features.
A set of color features and affective features extracted from music videos has also been proposed for music genre classification \cite{schindler2015audio}.  To capture different types of semantic information, they use various psychological or perceptive models. 

To match  videos and music, Chao et al. \cite{chao2011tunesensor} introduced a technique to recommend suitable music for photo albums and our approach  is similar. They utilized a cross-modal graph in which synsets of mood tags obtained from images and music are used as vertices, and relations between synsets are expressed as edges. 
Recently, Sasaki et al. \cite{sasaki2015affective} proposed a one-directional video-to-music recommendation by using the valence arousal plane. 
The approach most closely related to our study is that proposed by Wu et al. \cite{wu2016bridging}. To connect music and images, the authors used text from lyrics as a go-between media (i.e., metadata) and applied canonical correlation analysis \cite{shawe2004kernel}.

In summary, many approaches have been proposed to perform the task of associating music and visual modalities, and have shown remarkable performance,
However, most of the existing methods  use metadata (e.g., keywords, mood tags, and associated description) \cite{liem2012musesync, shah2014advisor, wu2016bridging} to associate music with  visual modalities. In contrast to these approaches, we directly connect items from the music and visual modalities using only the contents of each modality.

\begin{figure*}
	\centering
	\includegraphics[width=0.98\linewidth]{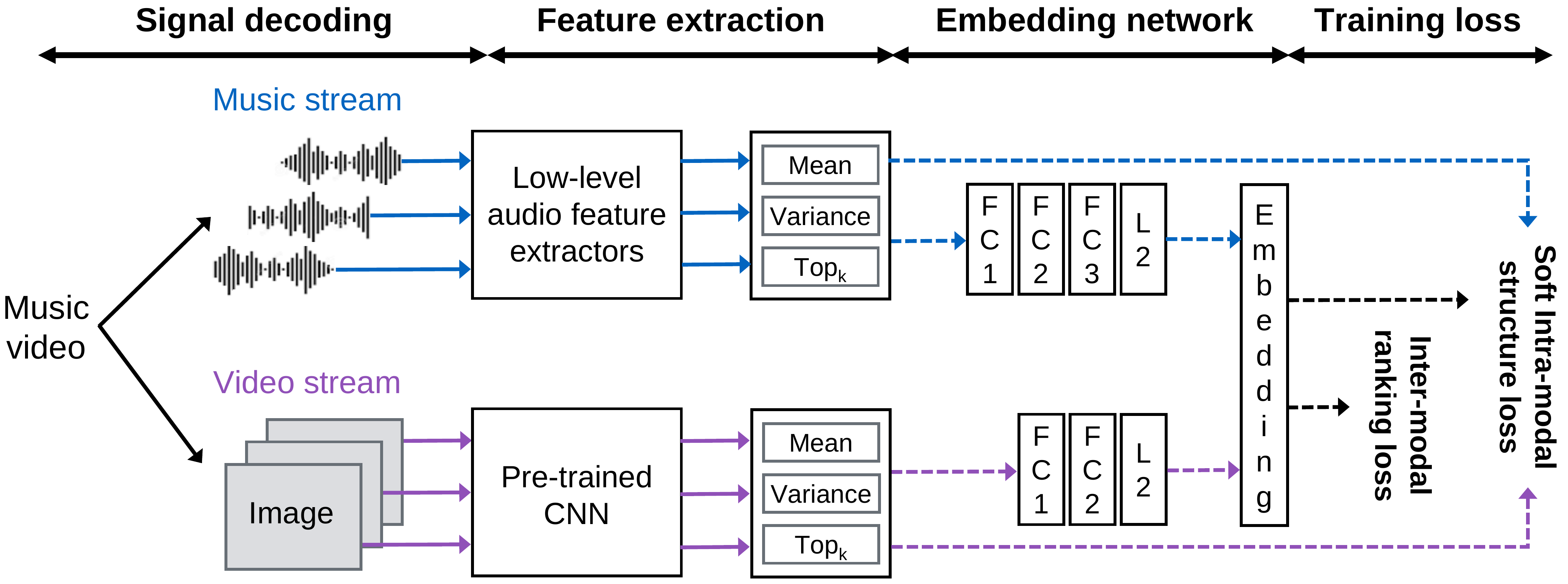}
	\caption{
Framework for the proposed method: Given a video and its associated music as input, we extract video features through a pre-trained CNN and extract music features through low-level audio feature extractors. For each modality, the features are then aggregated and fed into a two-stream neural network that is followed by an embedding layer. The whole embedding network is trained by two losses with different purposes: ranking loss for inter-modal relationship and our newly proposed loss for soft intra-modal structure. Each dotted arrow indicates a flow that can be trainable, while the solid arrows indicate flows that does not.
	}
	\label{fig:overallflow}

\end{figure*}

\subsection{Two-branch Neural Networks}
\label{sec:two_branch}

Over the past decade, considerable efforts have been devoted to study the relationship between different modalities.
In particular, recent studies have mostly focused on techniques for associating images with text encouraged by the promising results of deep learning techniques \cite{hong2017sspp, bae2016beyond, hong2017recognizing, han2016deep}.
From the viewpoint of architecture, our approach is similar to the existing two-branch neural networks for image\textendash text related tasks. 
In this section, we discuss the similarities and differences between our method and conventional approaches.

Traditionally, most studies investigating image\textendash text matching have been carried out on MSCOCO \cite{lin2014microsoft} and Flickr30K \cite{young2014image} in which each image in the datasets is described independently by five captions.
Early works based on two-branch neural networks have involved WSABIE \cite{weston2011wsabie} and DeVISE \cite{frome2013devise}. They commonly learn linear transformations of features from visual and textual modalities to the shared embedding space using a single-directional ranking loss. They apply a margin-based loss to an incorrect text annotation when it gets ranked higher than a correct one for describing an image. 
To better understand the relationship between visual and textual modalities, a few other works have proposed a bidirectional ranking loss  \cite{karpathy2015deep, karpathy2014deep, kiros2014unifying}.

In considering the intra-modal structure, image\textendash text embedding networks \cite{wang2016learning} and similarity networks \cite{wang2017learning} are most related to our method.
They propose a structure-preserving constraint that uses ground truth pair information within individual modality  resulting in an embedding space where the five captions corresponding to a particular image are close together.
Crucially, in contrast to these approaches \cite{wang2016learning, wang2017learning}, our approach does not use pair information within individual modality.
Instead, we take advantage of the relative distance relationships between the samples within the individual modality.
Our soft intra-modal structure constraint does not require intra-modal pair information, and it can be applied in more general situations where one-to-many or many-to-many cross-modal pair information is not available.

In addition to the techniques employed for image\textendash text matching, some works based on the two-branch neural network for music and visual modalities have  also  been proposed.
Among them, the work by Acar et al. \cite{acar2014understanding} is closely related to our approach. 
They presented a framework for the affective labeling of music videos in which higher-level representations are learned from low-level audio-visual features using a convolutional neural network (CNN). However, unlike our model, they use deep architecture only as a mid-level feature extractor for music and images and do not learn inter-modal relationships. 

\section{The Proposed Method}
Our goal is to design a model that infers the latent alignment between video and music as shown in Fig.~\ref{fig:overallflow}, enabling the retrieval of music that suits a given video and vice versa.

\subsection{Music Feature Extraction}\label{sec:music_feature}
To represent the music part, we  first decompose an audio signal into harmonic and percussive components, following previous studies  \cite{canadas2014percussive, choi2016convolutional}. We then apply log-amplitude scaling to each component to avoid numerical underflow. Next, we slice the components into shorter segments called local frames (or windowed excerpts) and extract multiple features from each component of each frame.

\textbf{Frame-level features.}
(1) Spectral features: The first type of audio features are derived from spectral analyses. In particular, we first apply the fast Fourier transform and the discrete wavelet transform to the windowed signal in each local frame. From the magnitude spectral results, we compute summary features including the spectral centroid, the spectral bandwidth, the spectral rolloff, and the first and second order polynomial features of a spectrogram \cite{fu2011survey}. 
(2) Mel-scale features: To extract more meaningful features, we compute the Mel-scale spectrogram of each frame as well as the Mel-frequency Cepstral Coefficients (MFCC). Further, to capture variations of timbre over time, we also use delta-MFCC features, which are the first- and second-order differences in MFCC features over time. 
(3) Chroma features: While Mel-scaled representations efficiently capture timbre, they provide poor resolution of pitches and pitch classes. To rectify this issue, we use chroma short-time Fourier transform \cite{khadkevich2013reassigned} as well as chroma energy normalized \cite{muller2011chroma}. 
(4) Etc.: We use the number of time domain zero-crossings as an audio feature in order to detect the amount of noise in the audio signal. We also use the root-mean-square energy for each frame. Table~\ref{table:audio_feature} summarizes the audio features utilized in our approach.

\begin{table}[b]
  \caption{Low-level audio features used in our approach.}
  \label{table:audio_feature}
   \begin{normalsize}
  \begin{tabular}{cl}
    \toprule
    Type & \qquad\qquad\qquad Audio features\\
    \midrule
   \multirow{2}{*} {Spectral} & spectral centroid, spectral bandwidth, \\
      &  spectral rolloff,  poly-feature ($1st$ and $2nd$)\\ \hline
     {Mel-scale} & MFCC, Mel-spectrogram, \\ &  delta-MFCC ($1st$ and $2nd$) \\\hline
      Chroma & chroma-cens, choroma-STFT \\ \hline
       Etc. &  zero-crossing, root-mean-square  energy \\ 
      \bottomrule
\end{tabular}
\end{normalsize}
\end{table}

\textbf{Music-level features.}
For each frame-level feature, we calculate three music-level features that capture properties of the entire music signal.  In particular, we calculate first- and second-order statistics (i.e., mean and variance) of each frame-level feature as well as maximum top $K$ ordinal statistics, which provide vectors consisting of the $K$ largest values for each dimension of frame-level features.
Finally, all the calculated music-level features are concatenated, and the result is passed from the feature extraction phase to the embedding neural network, as shown in the music-stream portion of Fig.~\ref{fig:overallflow}.


\subsection{Video Feature Extraction}\label{sec:video_feature}

\textbf{Frame-level features.}
Because our proposed HIMV-200K dataset contains a large number of videos, it is impractical to train the deep network from scratch. Instead, following the work in \cite{abu2016youtube}, we extract frame-level features from the individual frames using an Inception network \cite{szegedy2015going} trained on the large-scale ImageNet dataset \cite{deng2009imagenet}. We then apply a whitened principal component analysis (WPCA) so that the normalized features are approximately multivariate Gaussian with zero mean and identity covariance. As indicated by \cite{abu2016youtube}, this makes the gradient steps across the feature dimension independent, and as a result, the learning process for the embedding space, which we will run later, converges quickly, insensitive to changes in learning rates.

\textbf{Video-level features.}
Once frame-level features are obtained, we apply feature aggregation techniques and then perform concatenation as in the process of obtaining the music-level features. We then apply a global normalization process, which subtracts the mean of vectors from all the features, and we apply principal component analysis (PCA) \cite{jolliffe2002principal}. Finally, we perform L2 normalization to obtain video-level features. 

\subsection{Multimodal Embedding} \label{sec:embed}

The final step is to embed the separately extracted features of the heterogeneous music and video modalities into a shared embedding space. The embedding of music and video occurs via a two-branch neural network, with separate branches for each modality followed by a embedding layer. The whole network is trained by ``inter-modal ranking constraint'' and our newly proposed ``soft intra-modal structure constraint.''

\textbf{The two-branch neural network.}
The music and video features obtained via the processes in Sec.~\ref{sec:music_feature} and  Sec.~\ref{sec:video_feature} are fed into  a two-branch neural network in which each branch contains several fully connected (FC) layers with rectified linear unit (ReLU) nonlinearities. 
Recall that video features are extracted from a pre-trained deep architecture, whereas the music features are simply concatenated statistics of low-level audio features. To compensate for the relatively low-level audio features, we make the audio branch of the network deeper than the video branch as shown in Fig.~\ref{fig:overallflow}.
The final outputs of the two branches are  L2-normalized for convenient calculation of cosine similarity, which is used as distance metric in our approach.

\begin{figure}[]
\centering
\begin{subfigure}[c]{\linewidth}
\includegraphics[width=\linewidth]{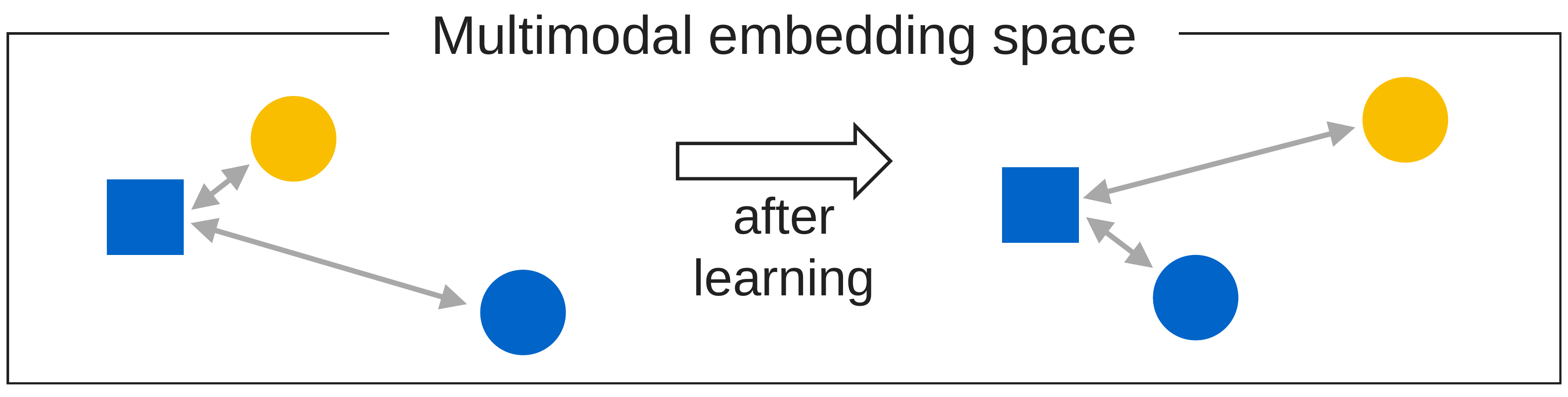}
\caption{\textbf{Embedding with inter-modal ranking constraint.} 
Videos and music with similar semantics are close together in embedding space.}
\label{fig:loss_concept_1}
\end{subfigure}
\begin{subfigure}[c]{\linewidth}
\includegraphics[width=\linewidth]{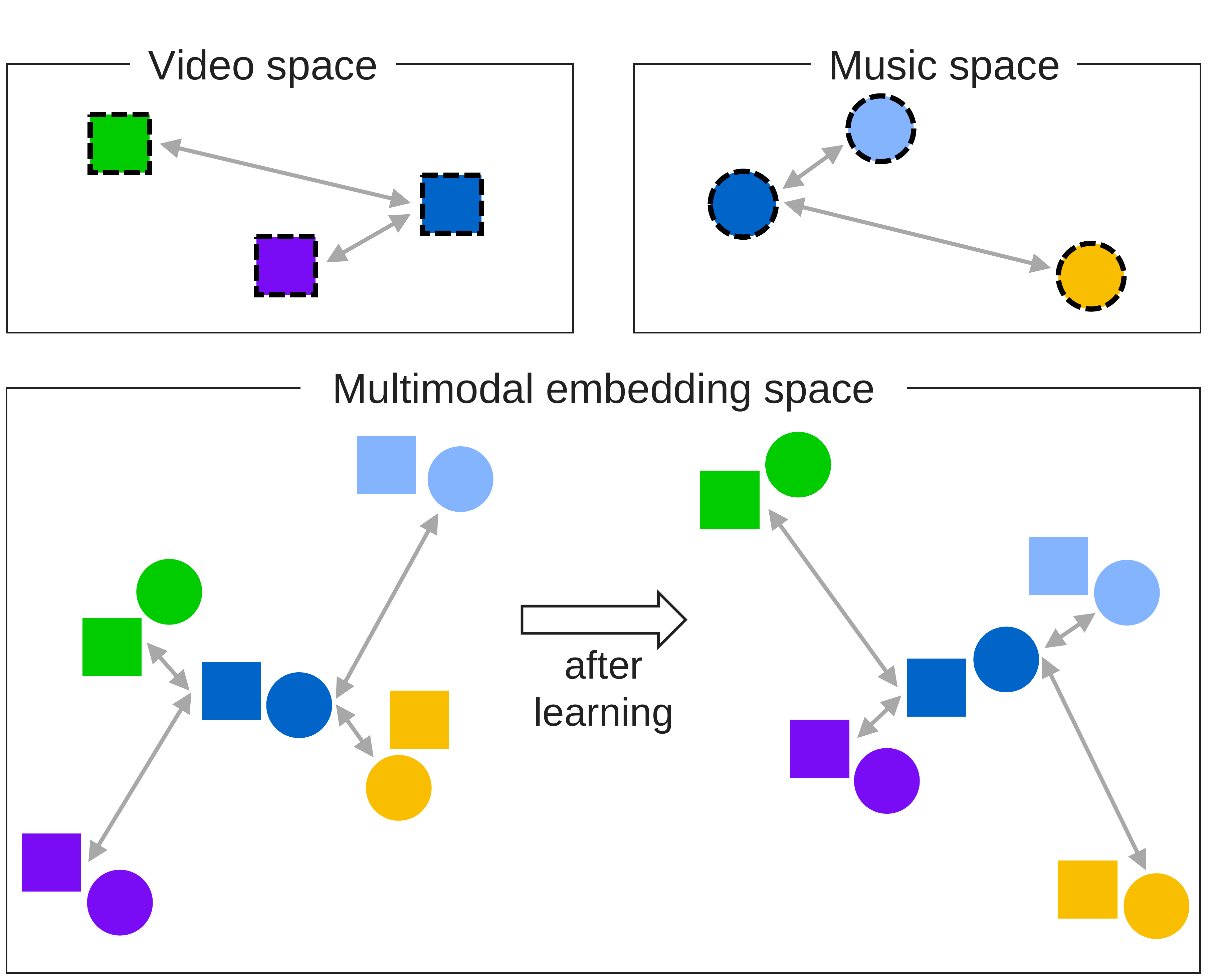}
\caption{\textbf{Embedding with (right) and without (left) soft intra-modal structure constraint.} 
If only inter-modal ranking constraint is used for embedding, the modality-specific characteristics  can be lost.
By leveraging the relative distance relationship between intra-modal samples (i.e.,  soft intra-modal structure) to be maintained in the embedding space, we can maintain the modality-specific characteristics in the embedding space.
}
\label{fig:loss_concept_2}
\end{subfigure}
\caption[Short]{
	Two constraints for multimodal embedding. 
	Identical shapes signify feature vectors obtained from the same modality. 
    The same color means that they come from the same music video. 
	Shapes with dotted outlines represent intra-modal samples before  embedding. 

}
\label{fig:constraint}
\end{figure}

\textbf{Inter-modal ranking constraint.} \label{sec:cross_rank}
For cross-modal matching, we wish to ensure that a video is closer to suitable music than unsuitable music, and vice versa. 
Inspired by the triplet ranking loss \cite{weinberger2009distance}, we used the inter-modal ranking constraint  in which  similar cross-modal input items are mapped to nearby feature vectors in the multimodal embedding space.
Assume triplets of items consisting of an arbitrary anchor, a positive cross-modal sample that is a ground truth pair item separated from the same music video, and a negative sample that is not paired with the anchor as shown  in Fig.~\ref{fig:constraint} (a).
The purpose of the inter-modal ranking constraint is to minimize the distance between an anchor and a positive sample while maximizing the distance between an anchor and a negative sample in the multimodal embedding space. 
The expression of the inter-modal ranking constraint of a given situation of visual input is as follows:

\begin{equation}\begin{aligned}\label{eq:inter-modal_1}
d(v_i, m_i) + e < d(v_i, m_j)
\end{aligned}\end{equation}
where, $v_i$ (anchor) and $m_i$ (positive sample) are the features in the multimodal embedding space obtained from the video and music of the $i$-th music video respectively. $m_j$ (negative sample) refers to the music  feature obtained from the $j$-th  music video. $d(v, m)$  indicate the distance (e.g., Euclidean distance) between two features in the embedding space, and $e$ indicates a margin constant.  
Analogously, given a music input, we set the inter-modal ranking constraint as follows: 
\begin{equation}\begin{aligned}\label{eq:inter-modal_2}
d(m_i, v_i) + e < d(m_i, v_j)
\end{aligned}\end{equation}

In the  triplet selection process, calculating  the loss for all possible triplets requires a considerable amount of computation.  Therefore, in our experiments, we select top $Q$ most violated cross-modal matches in each mini-batch. 
Concretely, we applied an inter-modal ranking constraint by selecting a maximum of  $Q$ violating negative matches that are closer to the positive pair (i.e., a ground truth video\textendash music pair) in the embedding space.
Theoretical guarantees of  this sampling scheme have been discussed in \cite{shaw2011learning}, though not only in the context of deep learning.

\textbf{Soft intra-modal structure constraint.}
If we only use the inter-modal ranking constraint for training the whole embedding network,  inherent characteristics within each modality (i.e., modality-specific characteristics) can be  lost as can be seen from the bottom left of Fig.~\ref{fig:constraint} (b).
Here, the modality-specific characteristics can include  rhythm, tempo, or timbre in music, and brightness, color, or texture in videos.
To solve the problem of collapsing the structure within each modality, we devised a novel soft intra-modal structure loss.

Suppose that three intra-modal features are extracted from different music videos using the process described  in Sec.~\ref{sec:music_feature} or Sec.~\ref{sec:video_feature}.
As these features have not yet been fed into the embedding network, they can maintain modality-specific characteristics.
In Fig.~\ref{fig:constraint} (b), from the perspective of video, blue is more similar to purple  than green.  From the perspective of music, blue is more similar to sky blue  than yellow.
By leveraging this relative distance relationship to remain in the embedding space, we can preserve modality-specific characteristics even after the embedding process.
The expression of our proposed soft intra-modal ranking constraint for a given situation of visual input is as follows: 
\begin{equation}\begin{aligned}\label{eq:intra-modal_1}
d(v_i, v_j) < d(v_i, v_k)\quad \text{if}\: d(\widetilde{v}_i, \widetilde{v}_j) < d(\widetilde{v}_i, \widetilde{v}_k)
\end{aligned}\end{equation}
where $v_i$, $v_j$, $v_k$ are the video features in multimodal space from the $i$-th, $j$-th, and $k$-th  music videos, respectively. $\widetilde{v}_i, \widetilde{v}_j, \widetilde{v}_k$  are also the video features from the $i$-th, $j$-th, and $k$-th  music videos, but they have not yet  passed through the embedding network.
Unlike the conventional  image\textendash text embedding approaches  \cite{wang2016learning, wang2017learning}  that leverage five captions  corresponding to one image (relatively hard structure or relationship), the relative distance relationship between samples is a soft relationship; thus, we do not use the margin constant when defining the soft intra-modal structure constraint for the music input as follows:
\begin{equation}\begin{aligned}\label{eq:intra-modal_2}
d(m_i, m_j) < d(m_i, m_k)\quad \text{if}\: d(\widetilde{m}_i, \widetilde{m}_j) < d(\widetilde{m}_i, \widetilde{m}_k)
\end{aligned}\end{equation}
where $m_i$, $m_j$, $m_k$ are the music features in multimodal space from the $i$-th, $j$-th, and $k$-th  music videos respectively and $\widetilde{m}_i, \widetilde{m}_j, \widetilde{m}_k$  are the corresponding music features before embedding.

\textbf{Embedding network loss.} 
Finally, we can combine the inter-modal ranking constraint  with our proposed soft intra-modal structure constraint  to construct an unified embedding network.
Given a mini-batch of $N$ music videos,  we can obtain $N$ pairs of embedded features of the form $(v_i,m_i)$, where $i$ ranges over the mini-batch. 
Here, $v_i$ and $m_i$ are the features from the video and music of the $i$-th music video, after passing through the two-stream embedding network. 
In this scenario, for the inter-modal ranking constraint, we can build two types of triplets $(v_i,m_i,m_j)$ and $(m_i,v_i,v_j)$, where $i\neq j$ are two different indices of music videos.
For the soft intra-modal structure constraint, we used two types of triplets  $(v_i,v_j,v_k)$ and $(m_i,m_j,m_k)$, where $i\neq j\neq k$ are three different indices of music videos.
Taking into consideration all these triplets, we set the network loss as



\begin{equation}\begin{aligned}\label{eq:total_loss}
L &= \lambda_1\sum^{}_{i\neq j}{max(0, v_i^T m_j -v_i^T m_i + e)} \\
  & + \lambda_2\sum^{}_{i\neq j}{max(0, m_i^T v_j  - m_i^T v_i + e)} \\
  	& + \lambda_3\sum^{}_{i\neq j\neq k}C_{ijk}(v){(v_i^T v_j  - v_i^T v_k)}\\
	& + \lambda_4\sum^{}_{i\neq j\neq k}C_{ijk}(m){(m_i^T m_j  - m_i^T m_k)}  \\ 
\end{aligned}\end{equation}
Here, $\lambda_1$ and, $\lambda_2$ balance the impact of inter-modal ranking loss from video-to-music and music-to-video matching. Recall that both video and music features are L2 normalized; therefore, as a distance measure, we use the negative value of the dot product between the video feature and the music feature. 
The impact of soft intra-modal structure loss within each modality was balanced by $\lambda_3$ and, $\lambda_4$, respectively. 
The function $C(\cdot)$ in  Eq.~\ref{eq:total_loss} is defined as follows: 
\begin{equation}\begin{aligned}\label{eq:coefficient}
C_{ijk}(x) =  sign(x_i^T x_k - x_i^T x_j) - sign(\widetilde{x}_i^T \widetilde{x}_k - \widetilde{x}_i^T \widetilde{x}_j)
\end{aligned}\end{equation}
where $x_i$, $x_j$, and $x_k$ are the trainable features in multimodal space and  $\widetilde{x}_i$, $\widetilde{x}_j$, and $\widetilde{x}_k$ are intra-modal features before passing through the embedding network.
$sign(x)$ is a function that returns one if  $x$ is a positive value, zero if $x$ is equal to zero, and $-1$ if $x$ is a negative value.
As we mentioned, the intra-modal structure is not rigid (or hard); thus, instead of using real distance values, we apply the $sign$ function when converting the  intra-modal structure constraint (\ref{eq:intra-modal_1}) and  (\ref{eq:intra-modal_2})  to our training objective (i.e., loss function).

\begin{figure*}[]
    \centering
    \begin{subfigure}[t]{0.32\textwidth}
        \centering
        \includegraphics[height=1.25in]{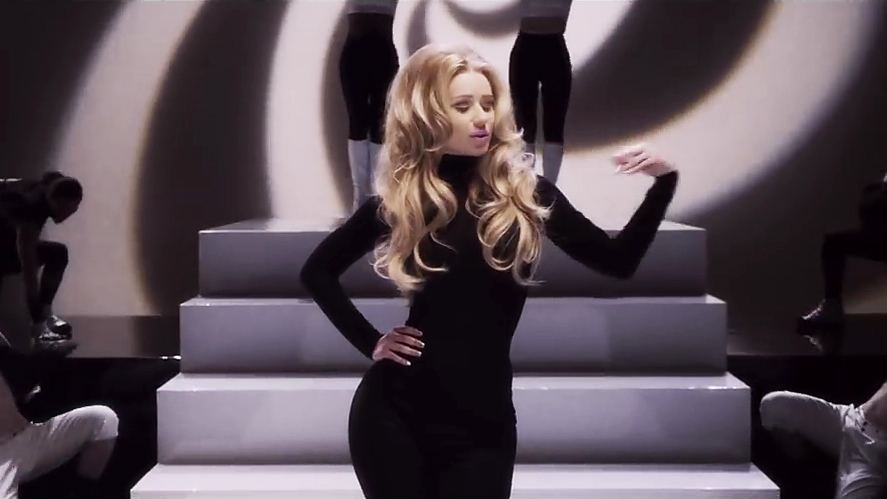}
        \caption{Official music video}
        \label{fig:eiffel}
    \end{subfigure}%
    ~ 
    \begin{subfigure}[t]{0.32\textwidth}
        \centering
        \includegraphics[height=1.25in]{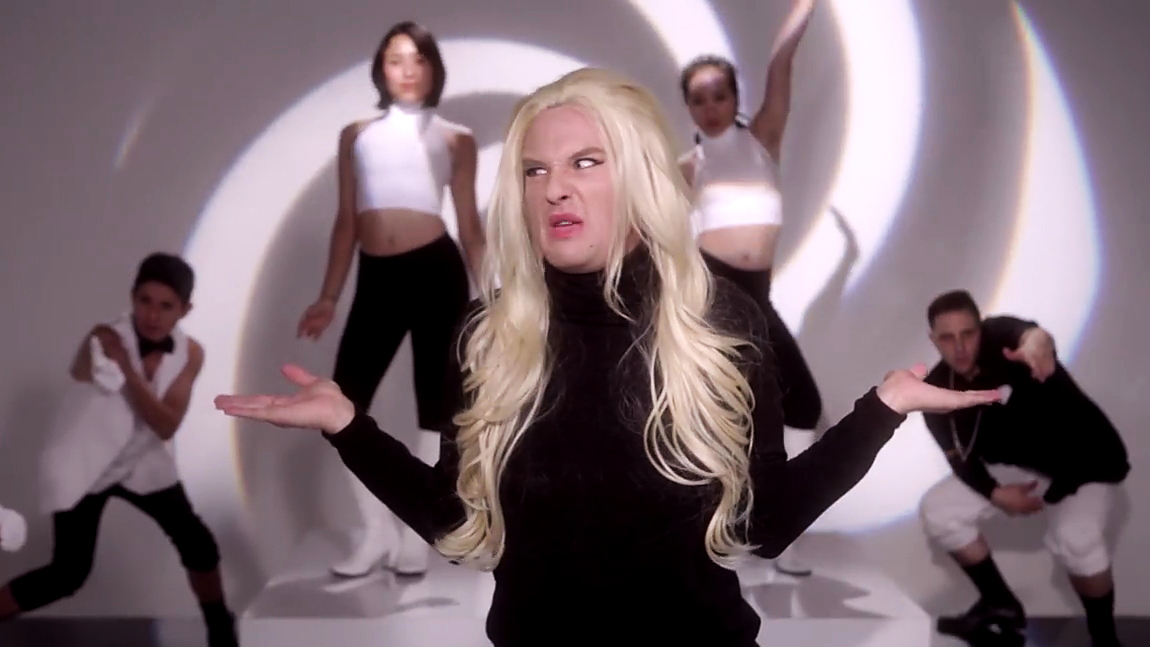}
        \caption{Parody music video}
        \label{fig:tajmahal}
    \end{subfigure}
    ~ 
    \begin{subfigure}[t]{0.32\textwidth}
            \centering
            \includegraphics[height=1.25in]{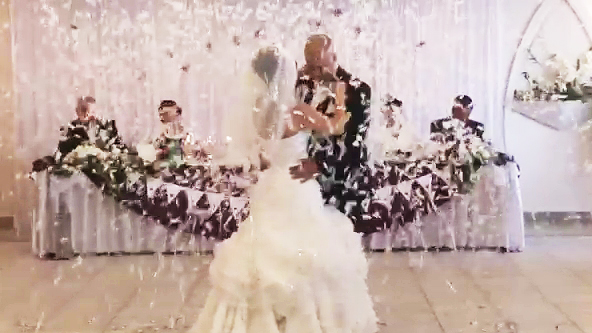}
            \caption{User-generated video}
            \label{fig:tajmahal}
        \end{subfigure}

    \caption{Sample images from HIMV-200K, video\textendash music pair datasets. }
    \label{fig:sample_mv}
\end{figure*}

\section{Dataset and Implementation Details}
In this section, we first introduce our large-scale  video\textendash music pair datasets that can be used for various tasks associated to music and visual modalities. We then present the details of our overall framework for CBVMR.

\subsection{Construction of the Dataset}
The practical challenge of CBVMR using data-driven learning such as deep learning is  the difficulty of obtaining well-matched video\textendash music pairs.
Until now, public datasets consisting of music and visual modality (e.g., images or videos)  have been rare, and unfortunately most of them are not accessible publicly as they are  copyrighted works.
Hence,  we  compiled a Hong\textendash Im Music\textendash Video 200K (HIMV-200K) benchmark dataset, composed of 200,500 video\textendash music.
We obtained these video\textendash music pairs  from the YouTube-8M,  a large-scale labeled video dataset that consists of millions of YouTube video IDs and associated labels \cite{abu2016youtube}.

All the videos in YouTube are annotated with a vocabulary of 4800 visual entities where the entities  include various activities (e.g., sports, games, hobbies), objects (e.g., autos, food, products), and events (e.g., concert, festival, wedding). 
Among the entire videos related to thousands of entities, we first downloaded videos related to ``music video,'' including official music videos, parody music videos, and user-generated videos with background music (Fig.~\ref{fig:sample_mv}).
Once all videos with a ``music video'' label were downloaded, they were separated into video and audio components using FFmpeg  \cite{bellard2012ffmpeg}. 
As a result, we obtained 205,000 video\textendash music pairs, and the sets for training, validation, and test comprised 200K, 4K, and 1K pairs, respectively.
To publicly distribute our HIMV-200K dataset without violating copyright, we provide the URLs of the YouTube videos under the  ``music video''  category and the  feature extraction code for both the videos and the music tracks  in our online repository. (https://github.com/csehong/VM-NET)

\subsection{Implementation Details}

As a music video usually lasts a few minutes, the audio signal separated from music video  may contain millions of samples given the high sampling frequency of greater than  10 kHz. Therefore, we trimmed the audio signals to 29.12 s at the center of the songs and downsample them from 22.05 kHz to 12 kHz following \cite{choi2016convolutional}. 
We then decomposed an audio signal into harmonic and percussive components and extracted numerous audio features frame by frame.
From this, we can obtain a vector of 380 dimensions for each frame. 
We calculated the first and second order statistics as well as maximum  top 1 ordinal statistics for each dimension of frame features and finally concatenate them for music-level features.

In video\textendash music retrieval, to represent video, we followed the implementation details in \cite{abu2016youtube}.
Each video was first decoded at 1 frame per second up to the first 360 s.
We extracted frame-level features of 2048 dimensions using Inception network \cite{szegedy2015going} and applied WPCA to reduce the feature dimensions to 1024. 
Given the frame-level features, we aggregated these features using the mean, standard-deviation, and top 5 ordinal statistics  followed by  global normalization.
As a result, a vector of 1024 dimensions was obtained as a video-level feature.

Finally, the details of our two-branch embedding network is as follows.
Given that high-level video features are extracted from the pre-trained CNN, we stacked only two FC layers with 2048 and 512 nodes for the video branch. On the other hand, after high-level music features are obtained from various low-level audio features, we stacked three FC layers with 2048, 1024, and 512 nodes. 
Finally, to solve the scale imbalance between music and image features, we apply batch normalization before the embedding loss layer in both video and audio part.
Our approach used the ADAM optimizer \cite{kingma2014adam} with learning rate 0.0003, and a dropout scheme with probability 0.9 is used. We selected a mini-batch size for training of 2,000 video\textendash music pairs.

\begin{table*}
\centering
  \caption{Retrieval results on HIMV-200K with respect to the key factors of VM-NET.  }
  \label{table:recall_key}
  \begin{normalsize}
  \begin{tabular}{|rr|r|r|r|r|r|r|}
   \hline
    \multicolumn{2}{|c}{\multirow{2}{*} {Parameter}}  & \multicolumn{3}{|c}{Music-to-video} & \multicolumn{3}{|c|}{Video-to-music} \\   \cline{3-8}
      &  & 	R@1	& R@10 & 	R@25& 	R@1	& R@10 & 	R@25\\  \hline
       \multicolumn{2}{|c|}{{Expected value of R@$K$}} &	  	0.1	& 1.0 & 	2.5 & 	0.1	& 1.0 & 	2.5\\ \hline
       \multirowcell{5}{Constraint weight \\($\lambda_{1}$, $\lambda_{2}$)}  
                 &	(1, 3)&	6.1	&12.3&	18.8	&5.8&	11.4&	17.3\\
            &	(1, 5)&	5.2	&12.0&	17.5 &	4.7	&11.6&	16.9\\
              &	(1, 1)&	8.4&	\textbf{19.3}&	28.6&	7.3	&17.6&	29.2\\
            &	(3, 1)&	8.5&	19.2	&\textbf{29.1}	&7.2&	\textbf{20.5}&	27.5\\
            &	(5, 1)&	\textbf{8.7}	&18.7&	27.4	&\textbf{7.7}&	18.0	&\textbf{29.3}\\ \hline     
       \multirowcell{5}{Number of layers \\ ($N_{music}$, $N_{video}$)}  &      (1, 1)&	4.3	&16.1&	25.6&	2.9	&15.0	&24.9 \\
 & (2, 2)&	7.5	&16.5&	27.3&	7.3	&16.0&	26.4\\
& (2, 3)&	7.4	&16.8&	26.4&	6.7	&16.2&	25.9\\
 &(3, 2)&	\textbf{8.5}	&\textbf{19.2}&	\textbf{29.1}&	7.2	&\textbf{20.5}&	\textbf{27.5}\\
 &(4, 2)&	8.2	&18.5&	27.3&	\textbf{7.5}	&16.5&	25.5\\ \hline 
        \multirowcell{5}{Number of top violations $Q$}    & 10 &	4.4 &	8.5	 &13.8	 &4.5 &	6.5	 &9.1 \\
             &50 &	8.0 &	17.3 &	25.7 &	7.4	 &16.5 &	25.4\\
             &100 &	8.5 &	19.2 &	29.1 &	7.2	 &20.5 &	27.5\\
             &500 &	8.4 &	21.2 &	31.7 &	\textbf{7.6}	 &19.4 &	\textbf{32.0}\\
             &1000 &	\textbf{9.1} &	\textbf{22.5} &	\textbf{32.6} &	\textbf{7.6}	 &\textbf{20.1} &	31.8\\\hline            
  \end{tabular}
  \end{normalsize}
\end{table*}

 \begin{table*}
 \centering
  \caption{Comparison of results on HIMV-200K using different methods.}
  \label{table:recall_method}
    \begin{normalsize}
  \begin{tabular}{|ll|r|r|r|r|r|r|}
   \hline
    \multicolumn{2}{|c}{\multirow{2}{*}{Method}}  & \multicolumn{3}{|c}{Music-to-video} & \multicolumn{3}{|c|}{Video-to-music} \\   \cline{3-8}
    &   &    R@1   & R@10 &    R@25&    R@1   & R@10 &    R@25\\ \hline
    \multirowcell{7}{Linear} & \multicolumn{1}{|c|}{PCA \cite{jolliffe2002principal}} & 0.0   &0.5   &2.1&   0.1&   1.1&   2.7\\ \cline{2-8}
      &\multicolumn{1}{|c|}{PLSSVD \cite{wegelin2000survey}}    & 0.7 &    5.9    & 13.0 &    0.8    & 6.8 &    14.1\\ \cline{2-8}
      &\multicolumn{1}{|c|}{PLSCA \cite{wegelin2000survey}}    & 0.8 &    6.3 &    13.7 &    0.4 &    7.1    & 14.1\\ \cline{2-8}   
      & \multicolumn{1}{|c|}{CCA \cite{hardoon2004canonical}} & 2.7 &   14.0&   25.8&   1.8   &14.3&   26.6       \\ \cline{2-8}
      & \multicolumn{1}{|c|}{Ours (one-directional)} & 4.2&   15.3&   24.4&   3.9   &15.7&  5 26.4\\ \cline{2-8}
      & \multicolumn{1}{|c|}{Ours (inter)}   & 4.3&   16.3&   27.0&   3.5&   15.3&   26.9\\ \cline{2-8}
       & \multicolumn{1}{|c|}{Ours (inter + intra)}   & 4.0&   16.7&   26.5&   3.4&   14.3&   25.9\\ \hline
      \multirowcell{4}{Nonlinear}   & \multicolumn{1}{|c|}{Ours (one-directional)}  & 6.5&   13.5&   19.9&   5.2   &12.0   &18.6\\ \cline{2-8}
           & \multicolumn{1}{|c|}{Ours (inter)}   &   \textbf{9.1} &	{22.5} &	{32.6} &	{7.6}	 &{20.1} &	31.8 \\ \cline{2-8}
         & \multicolumn{1}{|c|}{\textbf{Ours (inter + soft intra)}}   & {8.9}&   \textbf{25.2}&   \textbf{37.9}&   \textbf{8.2}  &\textbf{23.3}&   \textbf{35.7}\\ \cline{2-8}
       &\multicolumn{1}{|c|}{Ours (inter  + DANN \cite{ganin2016domain})} & 7.7&   19.7&   28.6&   {8.0}   &19.9   &29.8\\ \cline{2-8} 
       &\multicolumn{1}{|c|}{Ours (inter + soft intra + DANN \cite{ganin2016domain})} & 8.5&   24.0&   \textbf{37.9}&   {6.7}   &22.9   &35.4\\ \hline
       \multicolumn{8}{c}{inter: inter-modal ranking loss,\:\: soft intra: soft intra-modal structure loss} \\   
  \end{tabular}
    \end{normalsize}
\end{table*}

\section{Experimental Results}
To demonstrate the performance of our approach, we  conducted a quantitative evaluation as well as a subjective evaluation, and expect them to be the baselines for  less-mature video-to-music and music-to-video related tasks.

\subsection{The Recall@$K$ Metric}
\textbf{The protocol.}
Conventional video\textendash music related approaches has few standard protocols for evaluation, and even  most of them mainly performed subjective user evaluation.
To address this issue  we apply Recall@\$K, a standard protocol for cross-modal retrieval, especially in image\textendash text retrieval \cite{karpathy2015deep, wang2016learning} to the bidirectional CBVMR task. For a given value of $K$, it measures the percentage of queries in a test set for which at least one correct ground truth match was ranked among the top $K$ matches. For example, if we consider video queries that request suitable music, then Recall@10 tells us the percentage of video queries for which the top ten results include a ground truth music match.
To our knowledge, our study is the only work using Recall@$K$ to quantitatively evaluate the performance of bidirectional video\textendash music retrieval. The results we present in this section provide a baseline for future CBVMR tasks.

\textbf{Key factors.}
Motivated by \cite{karpathy2015deep, wang2016learning, wang2017learning}, we used a test set of 1,000 videos and their corresponding music tracks. Several key factors affected the performance of our model under the Recall@$K$ protocol as shown in Table~\ref{table:recall_key}. As a baseline, we provide the expected value for Recall@$K$ using random retrieval from 1,000 video\textendash music pairs.  All experimental results in this table  were obtained from our model using only inter-modal ranking loss (i.e.,  $\lambda_3$ and  $\lambda_4$ were set to zero)

The first section of the table shows the influence of the constraint weights. It is apparent that the performance can be generally improved by relatively giving more weight to $\lambda_1$ than $\lambda_2$. However, empirically,  we confirmed that setting $\lambda_1$ to five or more does not improve Recall@$K$. 

We also analyzed the results by changing the number of layers in the music and video branches of the embedding network. Using only one layer for each modality resulted in low performance, and it can be inferred that a deeper network is needed to generate an effective embedding. As the number of layers increases, the performance tends to be higher. We also confirmed that when the number of layers for the video part is more than three, the network weights do not converge in training. This appears to be due to degradation that often occurs in deep networks, as mentioned in \cite{he2016deep}. 

Finally, we also analyzed the results by changing the number of top violations, $Q$ (see Sec.~\ref{sec:cross_rank}). From the table, it is clear that there is a tendency for the performance to increase as $Q$ increases. In particular, the use of top 1000 violations results in a high retrieval rate of 9.1\% in the music-to-video task. We also performed experiments on the entire training mini-batch (i.e., $K$=2000) but could not confirm any additional performance improvements.

\begin{figure}[]
\centering
\begin{subfigure}[b]{\linewidth}
\includegraphics[width=\linewidth]{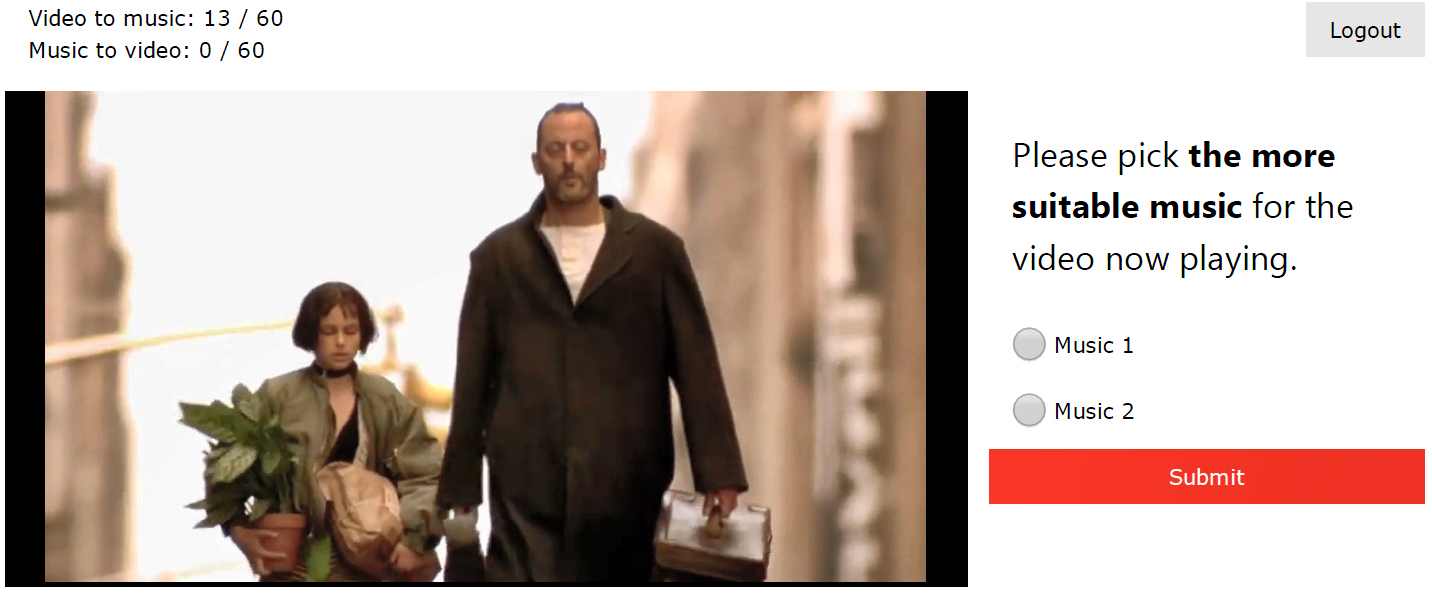}
\caption{Video-to-Music selection}
\label{fig:a2v_example}
\end{subfigure}
\begin{subfigure}[b]{\linewidth}
\includegraphics[width=1.02\linewidth]{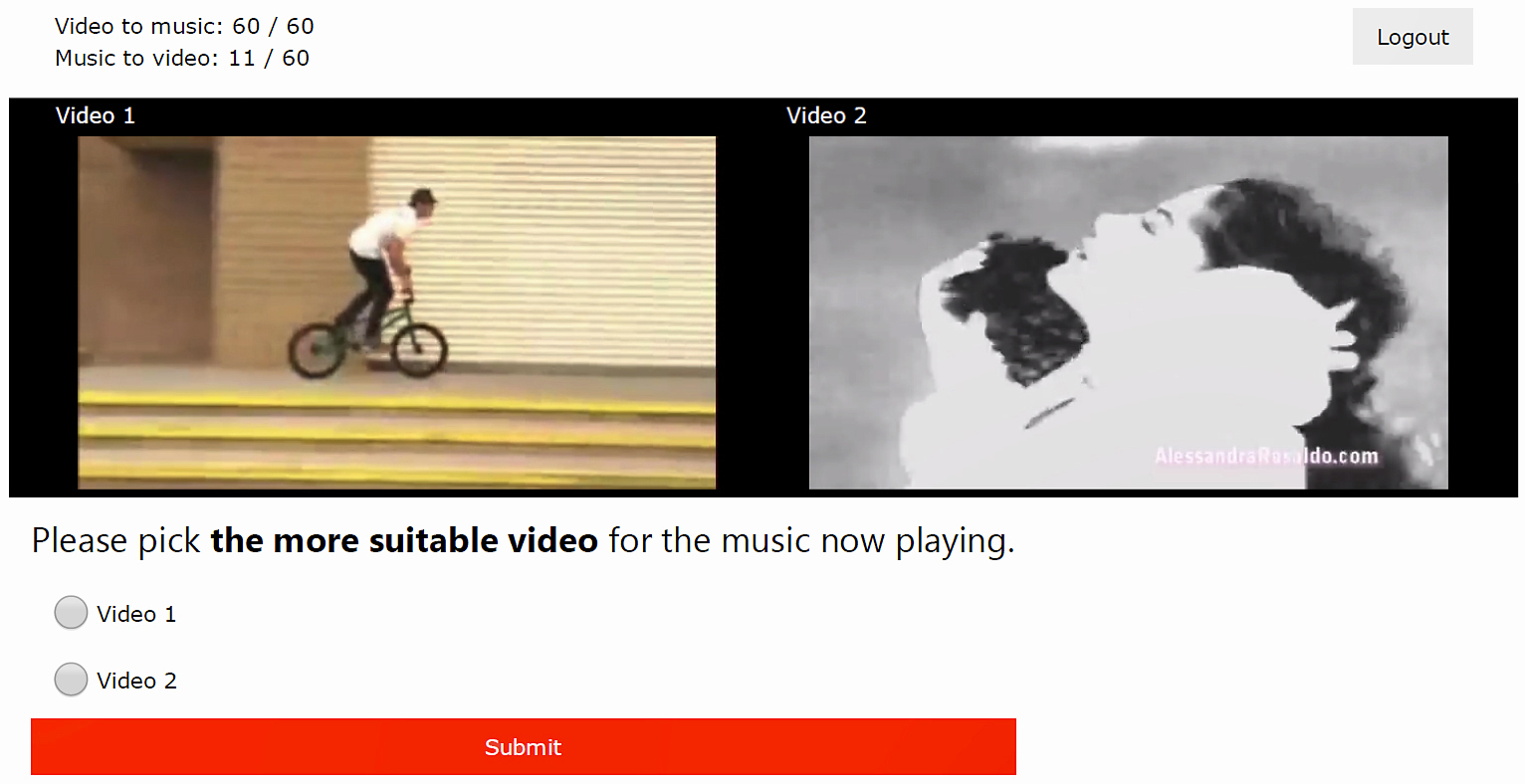}
\caption{Music-to-Video selection}
\label{fig:v2a_example}
\end{subfigure}
\caption[Short]{Screenshots from our human preference test}
\label{fig:test_system_example}
\end{figure}

\begin{figure*}
	\centering
	\includegraphics[width=\linewidth]{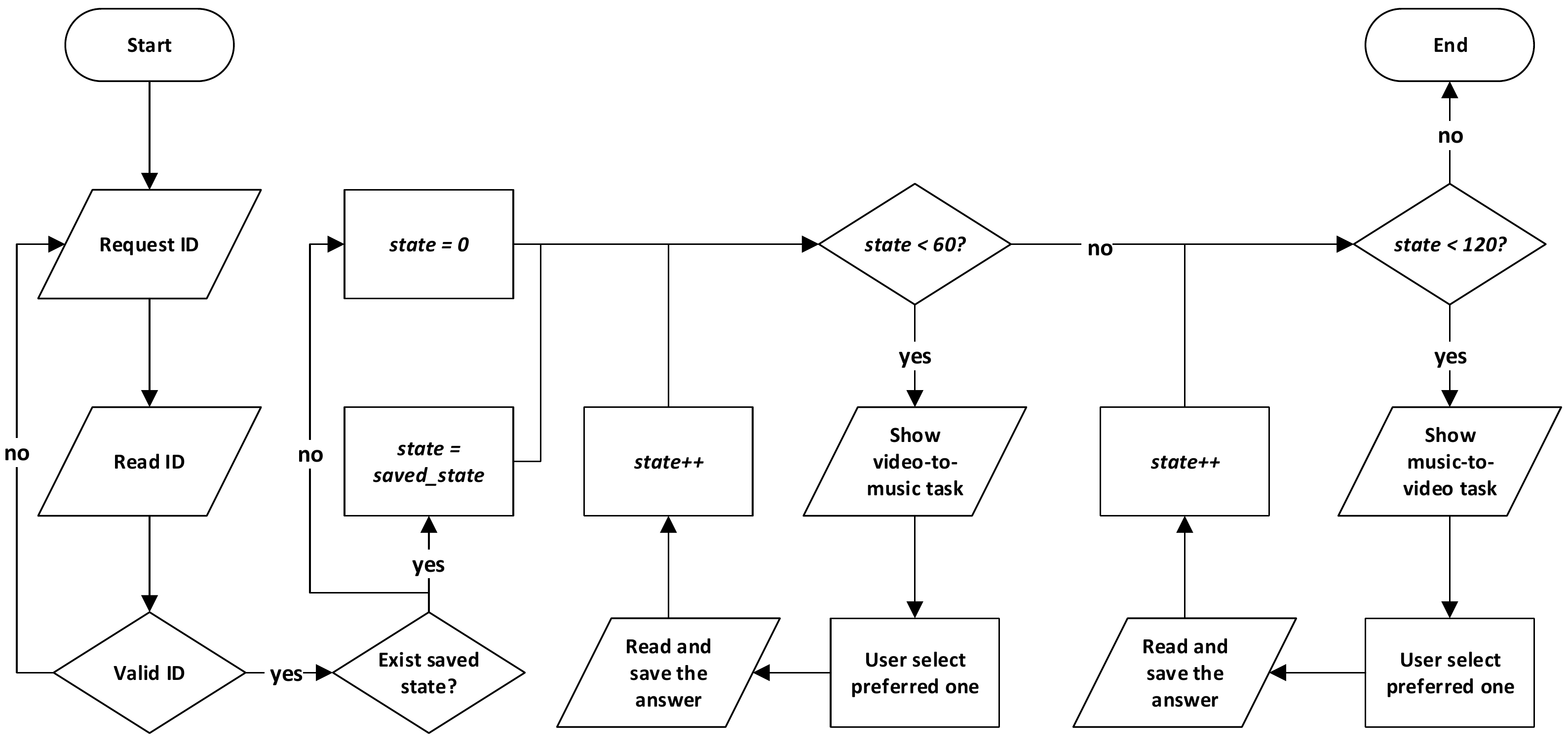}
	\caption{
Flowchart of the human preference test.
Fifty-three people participated in this test. We gave each subject 120 questions containing 60 video-to-music and 60 music-to-video selection tasks. For this task, we sent a unique ID  to every subject. When they had access to the address by web browsers on their computers, they performed video-to-music and music-to-video tasks sequentially. During the task, all progress was recorded into the server, so that one could resume the task even in case of the accidental shutdown. 
	}
	\label{fig:flowchart}

\end{figure*}

\textbf{Comparison with other methods.}
Table~\ref{table:recall_method} presents the  comparisons of our VM-NET with previous work. As a baseline, we present the result from linear models such as PCA \cite{jolliffe2002principal}  and techniques based on partial least squares (PLS) \cite{wegelin2000survey}. We also present the result of canonical correlation analysis (CCA) \cite{hardoon2004canonical}, which finds linear projections that maximize the correlation between projected vectors from the two modalities. 
To fairly compare the conventional linear models to our own using  bidirectional loss (i.e., inter-modal ranking loss), we evaluated the performance of our model without ReLU nonlinearities. We also present the results of our linear model using one-directional loss (i.e.,  $\lambda_1$, $\lambda_3$, and $\lambda_4$ were set to zero), and our model  is very similar to the WSABIE image\textendash text model \cite{weston2011wsabie}. 
Inspired by a technique \cite{park2016image} that applied the domain-adversarial neural network (DANN) \cite{ganin2016domain} to image\textendash text retrieval, we also applied DANN to our model.

From the experiment, it is clear that  obtaining a good retrieval performance with conventional linear models is difficult.
In nonlinear models, using bidirectional constraints instead of one-directional constraints improves the retrieval results by 2-3\% for R@1  and by a larger amount for the R@10 and R@25 protocols.
We also confirmed that applying DAAN to our model provides similar results, or even deteriorates the results, showing that the vanilla domain adaptation (DA) is not suitable for CBVMR tasks. 

As the matching criteria between video and music  are ambiguous than that of other cross-modal tasks such as image-to-text retrieval, the result of  R@10 and  R@25  may be more appropriate evaluation criteria than that of R@1 in CBVMR tasks. 
It is apparent from Table~\ref{table:recall_method} that using both inter-modal and intra-modal loss results in a significant performance improvement over using  only inter-modal loss, especially in the performance of R@10 and R@25.
This result suggests that the proposed soft intra-modal structure that preserves modality-specific characteristics is helpful for inferring the latent alignment between videos and music tracks.

\subsection{A Human Preference Test}

Applying Recall@$K$ to the relatively subjective task of cross-modal video\textendash music retrieval is one way to evaluate performance, but it might not be the most appropriate protocol. Therefore, we suggest and conduct a reasonable subjective evaluation protocol using humans. 

\textbf{Outline.}
We developed a web-based application in which test subjects were asked to pick one of two cross-modal items that the user thought was the best match to a given query (see Fig.~\ref{fig:test_system_example}). 
Fig. ~\ref{fig:flowchart} shows the overall flow diagram of the human preference test for video-to-music and music-to-video selection.

\begin{table}[]
\centering
\caption{Results of the human preference test, showing the percentage of subjects preferring one possible match to another. 
 The type of pairs, (G-R), (G-S), (S-R), were not informed to the subjects in this test. }
\label{table:user_test_result}
\setlength\tabcolsep{11pt} 
  \begin{normalsize}
\begin{tabular}{|c|c|c|c|c|} \hline
 Subject & Preference & V2M & M2V & Total \\ \hline 
\multirow{3}{*}{Human} &  G$>$R & 81.98 & 81.98 & 81.98\\
&G$>$S & 65.57 & 64.62 & 65.09\\
&S$>$R & 71.79 & 74.25 & 73.02\\ \hline 
Machine & G$>$R & 78.45 & 78.50 & 78.48\\\hline 
\multicolumn{5}{l}{\qquad\quad G: ground truth, R: random, S: search result.} \\
\multicolumn{5}{l}{\qquad\quad V2M: Video-to-Music, M2V: Music-to-Video.}
\end{tabular}
  \end{normalsize}
\end{table}

\textbf{Selection pairs.}
To analyze the users' subjectivity in detail, we used three different types of answer pairs. In some queries, the user was required to choose between a ground truth match and a randomly chosen match (\textbf{G-R}). Other queries required a choice between a ground truth match and a search match (\textbf{G-S}); still other queries had a choice between a search match and a random match (\textbf{S-R}). Here, ``ground truth'' indicates the cross-modal item extracted from the music video from which the query item was also extracted; ``random'' means a randomly selected item; and ``search'' indicates the closest cross-modal item to the query in the embedding space of our VM-NET model.

\begin{figure}[t]
\includegraphics[width=\linewidth]{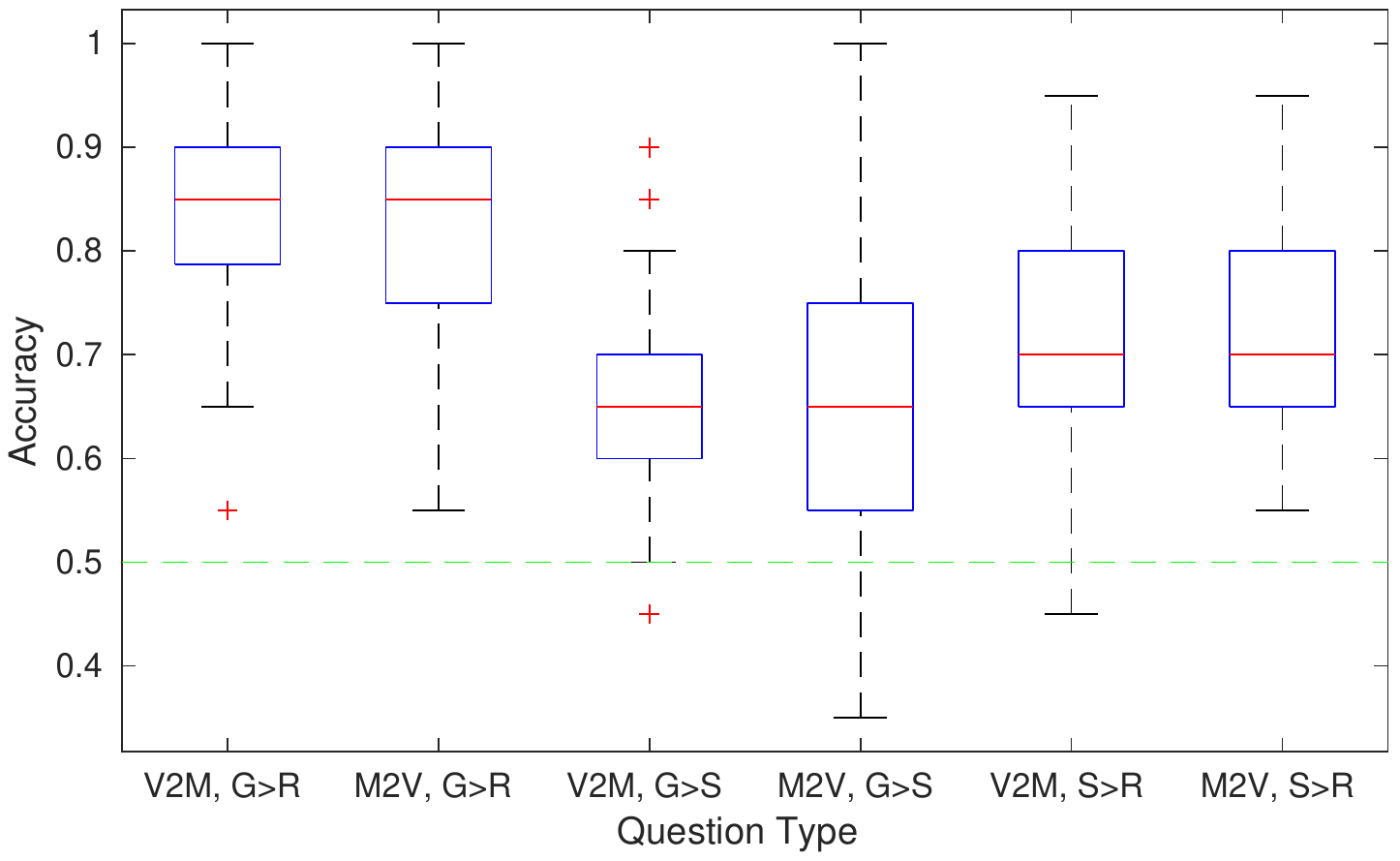}
\caption[Short]{Box plot showing the range of individual differences in preferences: The green line indicates the expected value if the user were to select randomly. The red lines indicate the median preference value taken across all subjects; the blue boxes represent the middle 50\% of subjects. The dotted black lines represent the full range of percentage values from all subjects, excluding outliers that are marked as red crosses.
}
\label{fig:result_box_plot}
\end{figure}

\begin{figure*}
\includegraphics[width=0.95\linewidth]{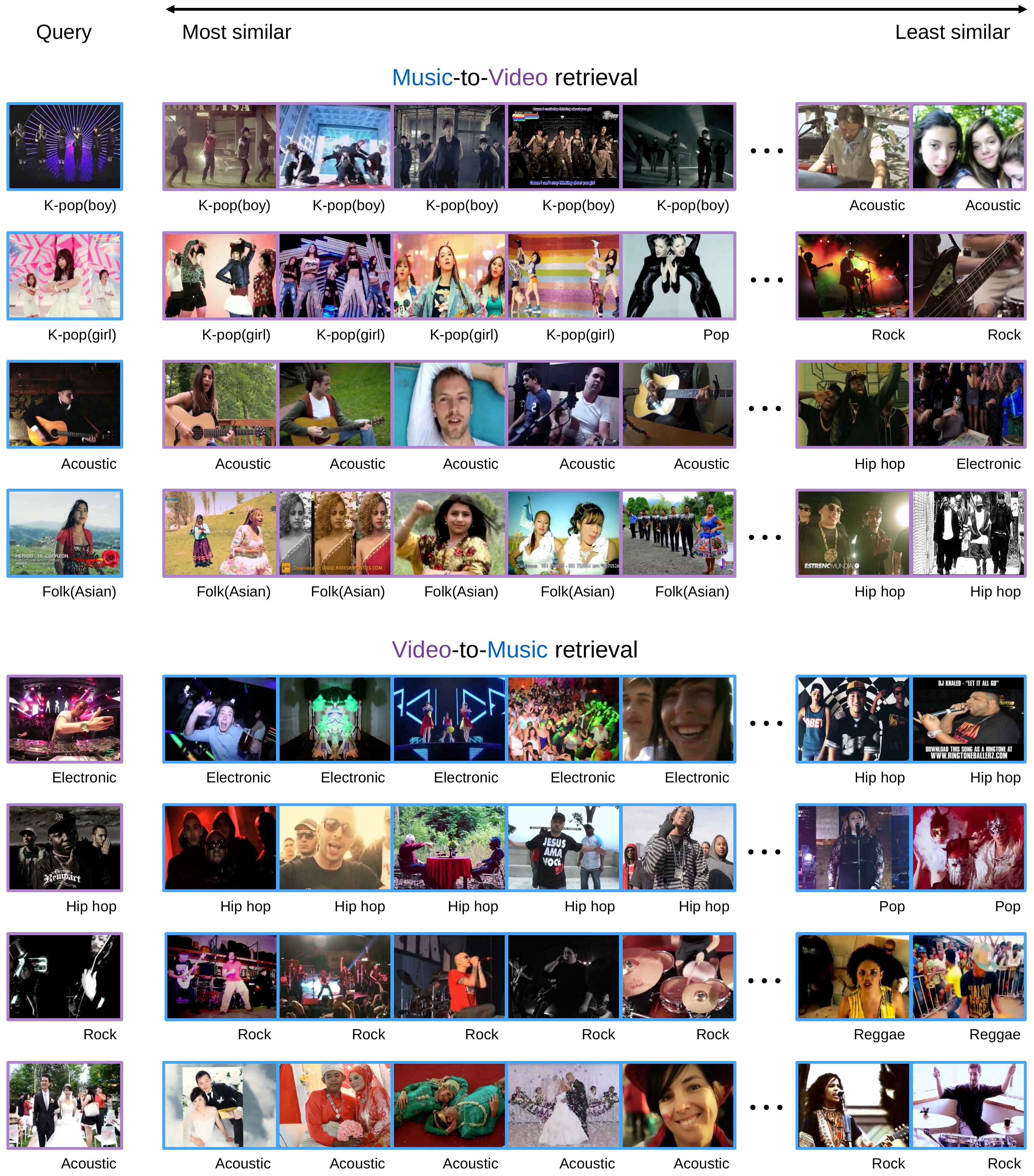}
\caption[Short]{
Bidirectional video\textendash music retrieval results for the HIMV-200 dataset: Note that the genre information written below each item was not used during the content-based retrieval, but that information aids in a qualitative assessment of the retrievals.
}
\label{fig:v2m_quality}
\end{figure*}

\begin{figure*}
\includegraphics[width=0.95\linewidth]{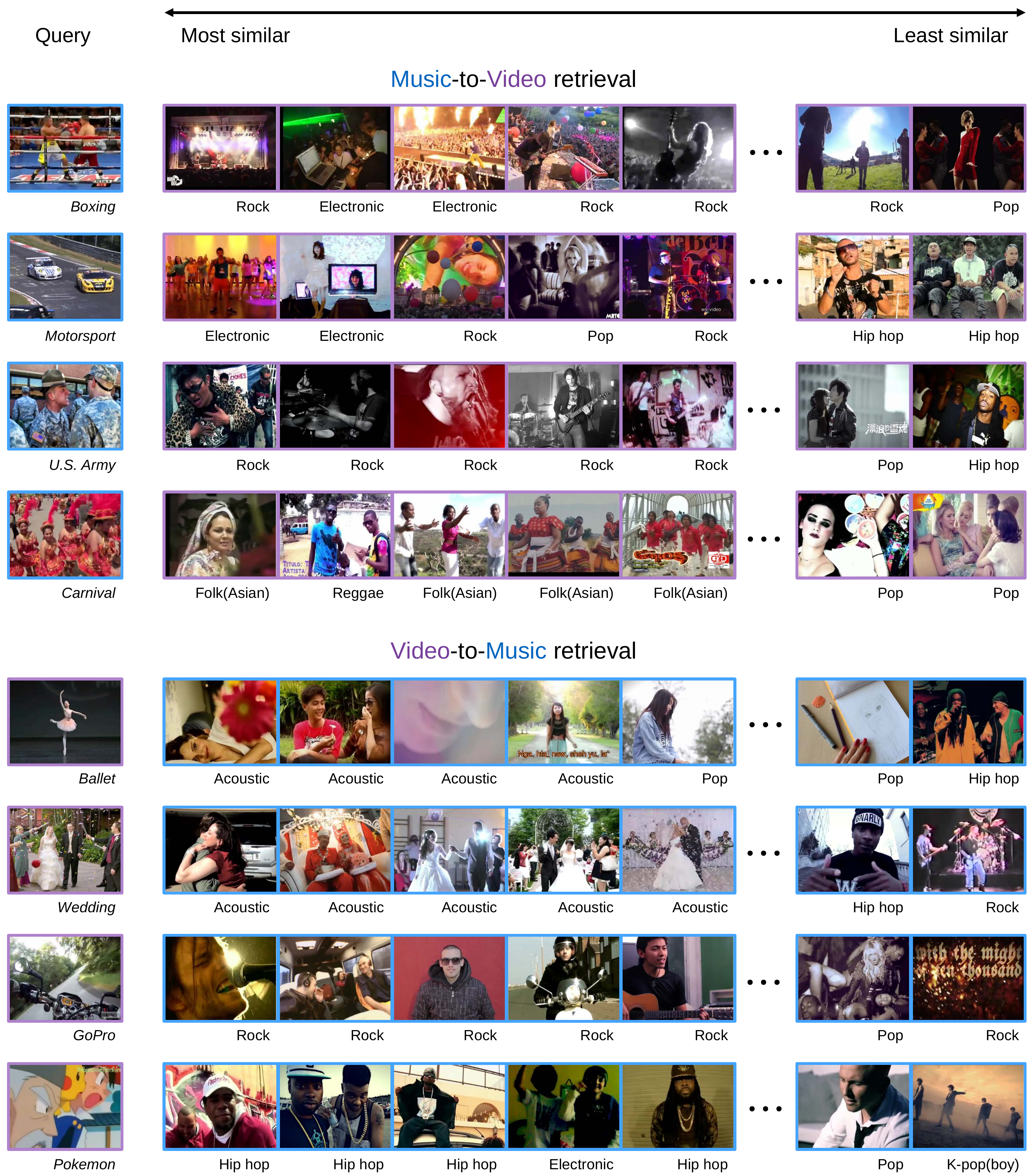}
\caption[Short]{Additional bidirectional video\textendash music retrieval results using  queries from videos and music that had been assigned by YouTube-8M to a general category:  Once again, the general category metadata written below each item was not used during the content-based retrieval, but it aids in a qualitative assessment.
}
\label{fig:v2m_quality_general}
\end{figure*}

\textbf{Analysis.}
Table~\ref{table:user_test_result} presents the results of the preference test, where each row (such as G$>$R) indicates the percent of answers where the subject preferred one match to another (such as preferring ground truth over random). Note that the human subjects' preference for G$>$R is over 80\%, which is much higher than the expected value of a random answer (i.e., 50\%). It follows that a human's preference for selecting video and music obtained from the same music video is fairly high, and this supports the validity of using video and music pairs from music videos for learning. 

Compared to G$>$R, G$>$S shows a clear decrease in human preference. In other words, humans had a very clear preference for ground truth over random matches, but that preference for ground truth decreases when the provided alternative is the search result from our model. This indicates that our model's results are better than the randomly selected results. That is, the retrieved results from our model are similar to ground truth items, which make it difficult for the subjects to choose. 
We can also see that the preference for S$>$R is over 70\%, which validates the effectiveness of our method in this subjective task. 
To show the  distribution of preferences in each task in more detail, we also present a box plot in Fig.~\ref{fig:result_box_plot}.

To compare the performance with human subjects and the machine (i.e., our model), we also experimented with what the machine preferred between ground truth items (G) and randomly selected items (R). In this experiment, we assumed that the machine always selected the item closer to the query in the embedding space. 
From  Table~\ref{table:user_test_result}, interestingly, we can see that the performance of our model in the G$>$R task is similar to human performance. This suggests a correlation between video and music regardless whether the judgment is by a  human or the machine.

\subsection{A Video\textendash Music Qualitative Experiment}

We also investigated the quality of the retrieval results for CBVMR tasks by using a test set comprising 1,000 video\textendash music pairs as shown in Fig.~\ref{fig:v2m_quality}. 
In the figure, pictures with the purple outlines represent the video items. Blue boxes represent music items. Because it is very difficult to express music or video itself only with figures, in both cases, we use just one key frame of the music video to represent its video item or the music item. Although the retrieval task was performed based only on content, we present in the figure a single piece of genre meta-information for each item to better convey the music or video content characteristics. Overall, the figure shows the results of each query in order from best match (on the left) to worst match (on the right). Even with the few examples shown in Fig.~\ref{fig:v2m_quality}, it is clear that the retrieved results accurately reflect characteristics of music or videos.

Finally, to demonstrate the flexibility of VM-NET, we performed additional experiments using video and music that had been assigned by YouTube-8M to a general category (e.g., ``boxing,'' ``motorsport,'' and ``U.S. army'') rather than just ``music video.''  (See Fig.~\ref{fig:v2m_quality_general}.)  Once again, the query results appear to be meaningful with respect to aspects such as gender and race of the characters. 

Finally, to directly visualize the retrieval performance of the proposed technique, we released a demo video online. (\href{https://github.com/csehong/VM-NET}{https://github.com/csehong/VM-NET})

\section{Conclusion}
In this paper, we introduced VM-NET, a two-branch deep network  that associates videos and music  considering inter- and intra-modal relationships.
We showed that inter-modal ranking loss widely used in other cross-modal matching is effective for the CBVMR task.
We also demonstrated that our newly proposed soft intra-modal structure preserving modality-specific characteristics significantly improves performance.
In the CBVMR task measured by Recall@$K$, our model outperformed conventional cross-modal approaches and this can be a 	quantitative baseline for subsequent CBVMR studies.
To evaluate how well our model relates the two cross-modal sources qualitatively, we also conducted a novel reasonable evaluation with human subjects. 
Finally, the quality of the bidirectional CBVMR results presented  suggests that the genre of music or gender or nationality of the  singer can be learned by our model.
In future work, we plan to expand our approach to a fully trainable architecture comprising low-level feature extraction as well as an embedding network.

\bibliographystyle{IEEEtran}
\bibliography{refs}

%
%
%
%
%




\end{document}